%% file: main.tex
\RequirePackage[svgnames]{xcolor}

\documentclass[11pt,letterpaper]{mystyle}

\usepackage[all]{hypcap}
\usepackage[svgnames]{xcolor}
\usepackage[comma,authoryear,compress]{natbib}
\usepackage{hyperref}[citecolor=lightblue]

\hypersetup{
    colorlinks = true,
    citecolor = {teal},
}

\usepackage{algorithm}
\usepackage{algorithmicx}
\usepackage{algpseudocode}
\usepackage{microtype}
\usepackage{graphicx}
\expandafter\def\csname ver@subfig.sty\endcsname{}
\usepackage{booktabs} %
\usepackage{float}
\usepackage{bigstrut}

\usepackage{amsmath}
\usepackage{amssymb}
\usepackage{mathtools}
\usepackage{amsthm}
\usepackage{mathrsfs}
\usepackage{nicefrac}
\usepackage{dsfont}
\usepackage{enumitem}
\usepackage{subcaption}
\usepackage{graphicx,subfig}
\usepackage{cleveref}
\usepackage{bxcoloremoji}
\usepackage{datetime2}
\usepackage{float}

\usepackage[utf8]{inputenc} %
\usepackage[T1]{fontenc}    %
\usepackage{hyperref}       %
\usepackage{url}            %
\usepackage{booktabs}       %
\usepackage{amsfonts}       %
\usepackage{nicefrac}       %
\usepackage{microtype}      %
\usepackage{graphicx}
\usepackage{subcaption} % Ensure subfigure or subtable environments are used if this package is included.
\usepackage{fdsymbol}
\usepackage{wrapfig}
\usepackage{lipsum}
\usepackage{enumitem}
\usepackage{stackengine}
\usepackage[font=small,labelfont=bf]{caption}
\usepackage{color}
\usepackage{adjustbox}
\usepackage{multirow, multicol}

\usepackage{rotating}
\usepackage{makecell}

\usepackage{tcolorbox}
\tcbuselibrary{listings,skins,breakable}
\usepackage{listings}
\usepackage{diagbox}

\input{macro}

\newtcolorbox{AIbox}[2][]{aibox,title=#2,#1}
\definecolor{lightgreen}{rgb}{0.22,0.70,0.30}%
\definecolor{Gray}{gray}{0.95}
\definecolor{Cornsilk}{rgb}{1.0, 0.97, 0.86}

\definecolor{lightblue}{HTML}{0064E0}
\definecolor{fg}{HTML}{1C2B33}
\definecolor{bg}{HTML}{F1F4F7}

\newcommand{\name}{\textsc{TALENT}}

\usepackage[justification=centering]{caption}

\usepackage{amsmath}

\usepackage[all]{hypcap}

\title{\LARGE TALENT: Table VQA via Augmented Language-Enhanced Natural-text Transcription}

\runningtitle{Table VQA via Augmented Language-Enhanced Natural-text Transcription}

\author[1]{Yutong Guo}
\author[2]{Wanying Wang}
\author[2]{Yue Wu}
\author[3]{Zichen Miao}
\author[4]{Haoyu Wang}

\affil[1]{Johns Hopkins University}
\affil[2]{Independent Researcher}
\affil[3]{Purdue University}
\affil[4]{University at Albany}

\begin{document}

\input{sections/abstract}
\maketitle
\vspace{3mm}
\input{sections/introduction}
\input{sections/relatedwork}
\input{sections/method}
\input{sections/experiments}
% \clearpage
\input{sections/conclusion}
\clearpage
\bibliography{main}

\end{document}

%% file: macro.tex
\newcommand{\x}{\mathbf{x}}

\definecolor{blanchedalmond}{rgb}{1.0, 0.92, 0.8}
\definecolor{carmine}{rgb}{0.59, 0.0, 0.09}
\definecolor{lightblue}{rgb}{0.22,0.45,0.70}%

\renewcommand{\mathbf}{\boldsymbol}

\makeatletter
\def\Ddots{\mathinner{\mkern1mu\raise\p@
\vbox{\kern7\p@\hbox{.}}\mkern2mu
\raise4\p@\hbox{.}\mkern2mu\raise7\p@\hbox{.}\mkern1mu}}
\makeatother

\definecolor{amaranth}{rgb}{0.9, 0.17, 0.31}
\definecolor{antiquebrass}{rgb}{0.8, 0.58, 0.46}
\definecolor{antiquefuchsia}{rgb}{0.57, 0.36, 0.51}
\definecolor{chromeyellow}{rgb}{0.31, 0.47, 0.26}

\newcommand{\github}{\raisebox{-1.5pt}{\includegraphics[height=1.05em]{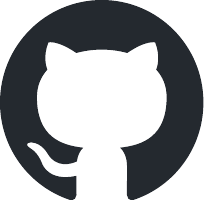}}}

%% file: sections/abstract.tex
\begin{abstract}
Table Visual Question Answering (Table VQA) is typically addressed by large vision-language models (VLMs). While such models can answer directly from images, they often miss fine-grained details unless scaled to very large sizes, which are computationally prohibitive, especially for mobile deployment. A lighter alternative is to have a small VLM perform OCR and then use a large language model (LLM) to reason over structured outputs such as Markdown tables. However, these representations are not naturally optimized for LLMs and still introduce substantial errors. We propose {\name} (Table VQA via Augmented Language-Enhanced Natural-text Transcription), a lightweight framework that leverages dual representations of tables. {\name} prompts a small VLM to produce both OCR text and natural language narration, then combines them with the question for reasoning by an LLM. This reframes Table VQA as an LLM-centric multimodal reasoning task, where the VLM serves as a perception–narration module rather than a monolithic solver. Additionally, we construct ReTabVQA, a more challenging Table VQA dataset requiring multi-step quantitative reasoning over table images. Experiments show that {\name} enables a small VLM–LLM combination to match or surpass a single large VLM at significantly lower computational cost on both public datasets and ReTabVQA.
\vspace{2mm}


\vspace{5mm}

\coloremojicode{1F4C5} \textbf{Date}: \today

% \coloremojicode{1F3E0} \textbf{Projects}: \href{https://wangrongsheng.github.io}{https://wangrongsheng.github.io}

\github{} \textbf{Code Repository}: \href{https://github.com/Melodramma080727/TALENT-Table-VQA-via-Augmented-Language-Enhanced-Natural-text-Transcription}{https://github.com/Melodramma080727/TALENT-Table-VQA-via-Augmented-Language-Enhanced-Natural-text-Transcription}

\coloremojicode{1F4DA} \textbf{Datasets}: \href{https://huggingface.co/datasets/yguo86/ReTabVQA }{https://huggingface.co/datasets/yguo86/ReTabVQA }

\coloremojicode{1F4E7} \textbf{Contact}: \href{mailto:yguo113@jh.edu}{yguo113@jh.edu};\href{mailto:hwang28@albany.edu}{hwang28@albany.edu}

\end{abstract}

%% file: sections/introduction.tex
\vspace{-4mm}
\section{Introduction}
\label{sec:intro}
\vspace{-1mm}
Table-based Visual Question Answering (Table VQA) has emerged as a crucial subfield of multimodal reasoning, requiring the recognition of visual tabular structures as well as natural language understanding and logical inference~\cite{kim2024tablevqabenchvisualquestionanswering}. Its importance is underscored by diverse real-world applications such as financial data analysis~\cite{srivastava2025enhancingfinancialvqavision}, medical report interpretation~\cite{zhang2024pmcvqavisualinstructiontuning}, education, and automated form processing~\cite{10.1007/978-3-031-70533-5_24}. On mobile and edge devices such as smartphones and tablets, users increasingly expect seamless and efficient interaction with structured data~\cite{10.1145/3534619}. For example, they may query invoices, exam scores, or financial statements using natural language. These scenarios highlight the growing demand for Table VQA systems that offer both accuracy and efficiency, enabling reliable and real-time interaction with tabular data in environments with limited resources~\cite{10.1145/3728635}.

To meet the growing demand for Table VQA on resource-constrained devices, recent methods have increasingly adopted vision-language models (VLMs) to process both visual and textual signals ~\cite{sinha2024guidingvisionlanguagemodelselection,kim2024tablevqabenchvisualquestionanswering}. These approaches generally fall into two paradigms. The first uses large, end-to-end VLMs to directly answer questions from table images~\cite{kim2024tablevqabenchvisualquestionanswering}. The second employs a VLM to perform Optical Character Recognition (OCR), then feeds the extracted content—often structured as Markdown or HTML—into a VLM or an LLM for reasoning~\cite{Patin2025ocr}.

\begin{figure}[h!]
% \vspace{-1em}
    \centering
\includegraphics[width=0.6\columnwidth]{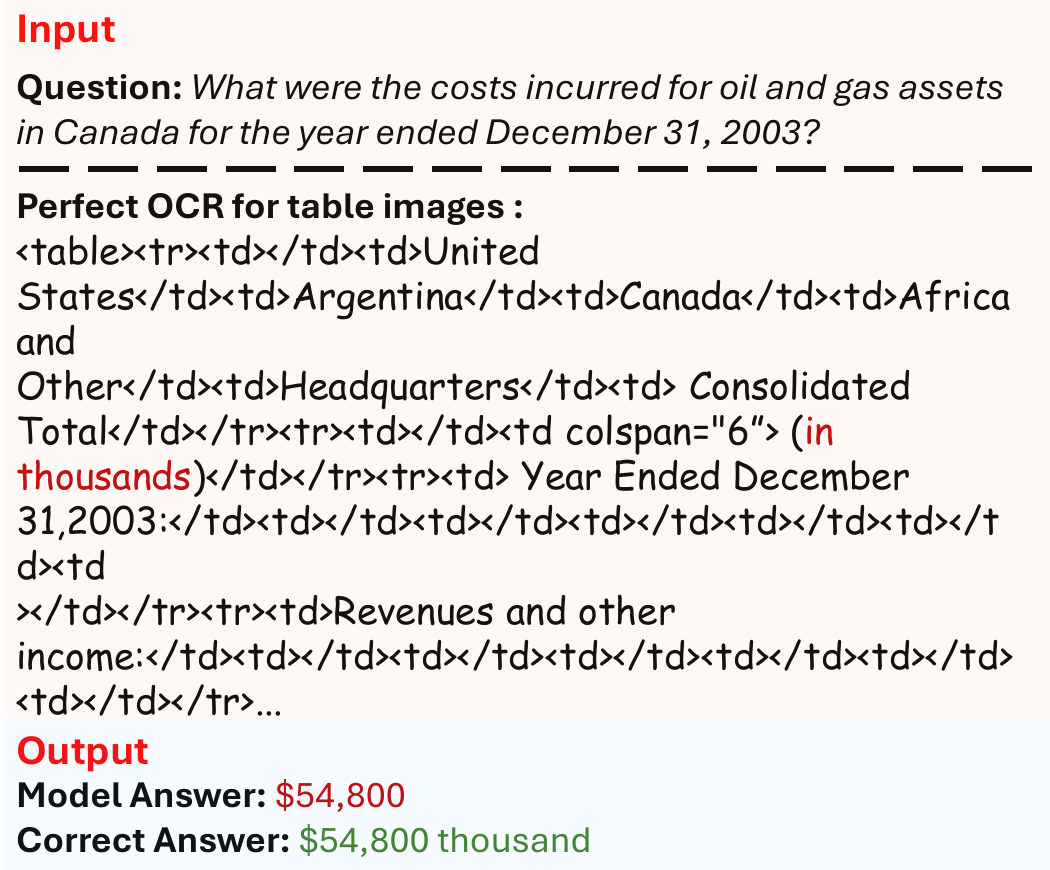}
    \caption{The critical dilemma in Table VQA. Existing methods often result in failure (red) by being either too computationally heavy or losing semantic context. Our goal is to achieve an accurate and efficient solution (green).}
    \label{fig:critical_gap}
    % \vspace{-1em}
\end{figure}

However, both paradigms exhibit notable limitations. End-to-end VLMs frequently overlook fine-grained visual details unless scaled to computationally demanding model sizes, e.g., VLMs with more than 70B parameters, thereby limiting their applicability in resource-constrained environments such as mobile or edge devices~\cite{10.1145/3534619, rashid2024tinyvqacompactmultimodaldeep}. In contrast, the OCR-to-LLM pipeline, while more lightweight, introduces alternative challenges. The special structured table representations, such as Markdown or HTML, are not inherently aligned with the reasoning capabilities of VLMs or LLMs, prompting efforts to decouple visual understanding from language-based inference~\cite{Patin2025ocr}. Moreover, the transformation from table images to structured formats can result in the loss of essential semantic cues embedded in the layout, including merged cells and header-to-cell relationships, as underscored by prior work on table structure recognition~\cite{raja2020tablestructurerecognitionusing}. Our preliminary analysis further indicates that even with perfect OCR, downstream LLMs frequently produce erroneous answers due to this contextual information loss. As illustrated in Figure~\ref{fig:critical_gap}, this leaves a critical gap in the current landscape: existing approaches are either too computationally intensive to deploy in real-world scenarios, or they are efficient but at the expense of answer accuracy.
% These findings suggest a critical gap in the current landscape: existing approaches are either computationally intensive and thus impractical for real-world deployment, or efficient but at the expense of answer accuracy.

To address these challenges, we propose {\name} (Table VQA via Augmented Language-Enhanced Natural-text Transcription), a framework that rethinks the role of VLMs in table reasoning. Instead of relying on oversized end-to-end VLMs or brittle structured representations, TALENT leverages the VLM as a perception–narration module that generates two complementary views of a table: (1) OCR spans providing precise symbolic content, and (2) natural language narration describing the table’s structure, headers, and salient values. These dual representations, combined with the user question, are passed to an LLM for reasoning. By aligning symbolic precision with semantic richness, TALENT enables the LLM to act as the central reasoning engine while using only small- or medium-sized models. This LLM-centric design reduces computational overhead and improves robustness to OCR noise and layout complexity, offering an efficient, deployable solution for Table VQA.

To rigorously evaluate multi-step reasoning and quantitative understanding, we construct a manually curated benchmark called the \textbf{ReTabVQA} Dataset (Reasoning Table VQA). Existing benchmarks such as TableVQA-Bench primarily emphasize straightforward information recognition and extraction from table images, which fails to assess whether a model comprehensively understands the table content. To address this limitation, ReTabVQA introduces compositional and numerically grounded questions that require multi-hop reasoning and quantitative synthesis across table entries, offering a more realistic and challenging evaluation setting.

The main contributions of this paper are summarized as follows: 1)~We propose {\name}, a new paradigm that reframes Table VQA as LLM-centric multimodal reasoning task, while leveraging the VLM as a perception–narration module to produce dual representations from table images. 2)~{\name} shows that collaboration between a small VLM and LLM can rival or surpass a single large VLM, achieving competitive accuracy with far fewer parameters—suitable for mobile and edge deployment. 3)~Experiments confirm the effectiveness of dual representations and the robustness of {\name} under complex table layouts. 4)~We contribute a new benchmark \textbf{ReTabVQA} for complex reasoning in Table VQA, containing 120 multi-step questions based on the TableVQA-Bench, enabling evaluation of VLM capabilities in challenging scenarios.

%% file: sections/relatedwork.tex
\section{Related Work}
\label{sec:related_work}

\begin{figure}[!t]
    \centering
    \includegraphics[width=0.75\textwidth]{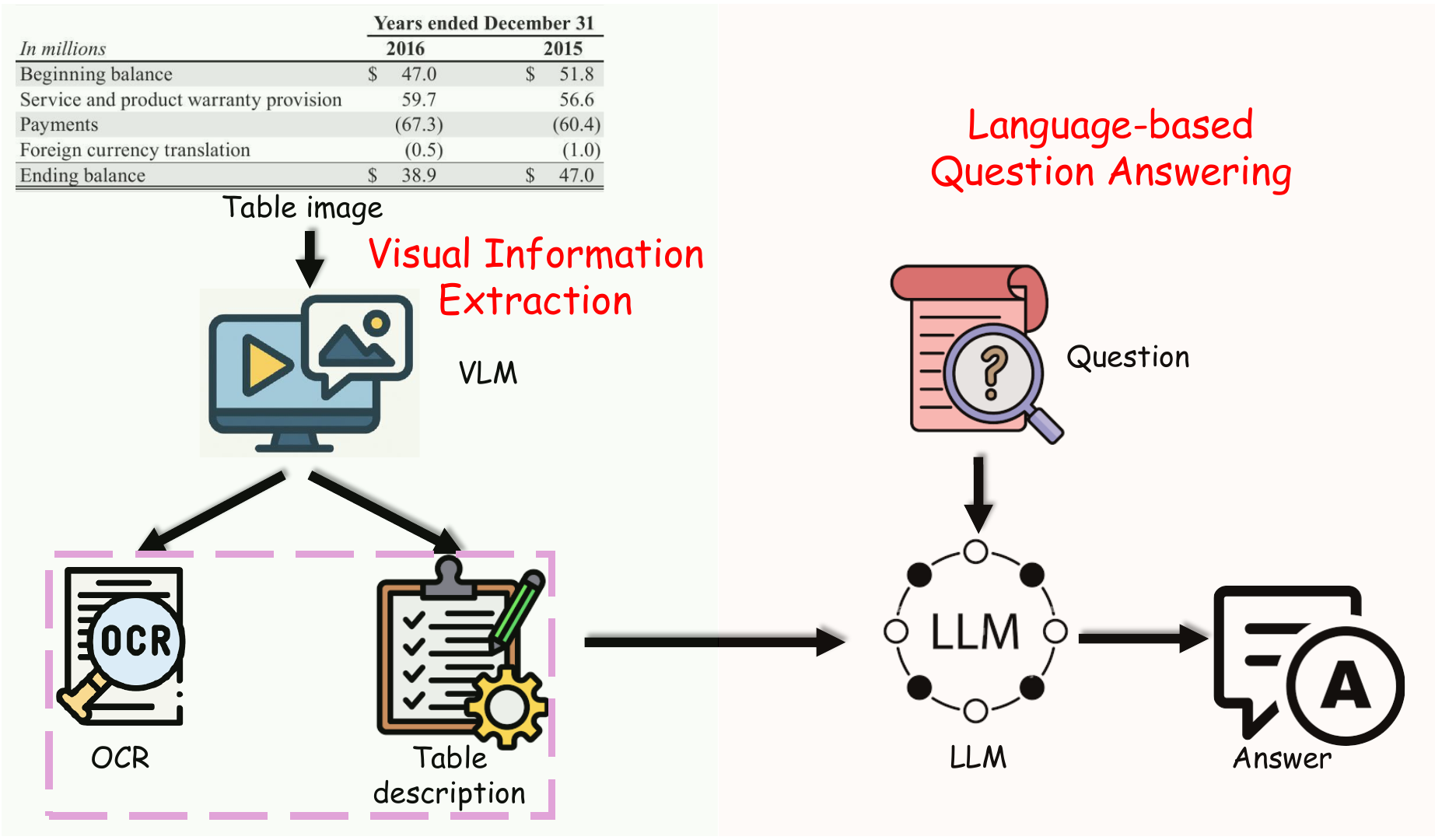}
    \caption{Overall framework of {\name}. {\name} redefines Table VQA by leveraging the VLM as a perception–narration module that generates both symbolic OCR spans and natural-language descriptions of tables. These dual representations, combined with the user question, are processed by an LLM as the central reasoning engine, enabling efficient and robust multimodal table understanding.
    }
    \label{fig:methodology}
    % \vspace{-1.5em}
\end{figure}

\subsection{End-to-End VLM Approaches in Table VQA}
The dominant paradigm in Table Visual Question Answering (Table VQA) has been the use of large, end-to-end Vision-Language Models (VLMs), such as LLaVA-1.5 and other instruction-tuned models~\cite{liu2024improvedbaselinesvisualinstruction,zhou2025enhancing,lompo2025opendomain,khan2023seltqa}. These models are trained to directly process a table image and a question to generate an answer~\cite{fang2024largelanguagemodelsllmstabular,hu2025llmtkie}. While they have demonstrated strong performance on various benchmarks, they face significant challenges. A primary limitation is their high computational cost, which makes them impractical for deployment on resource-constrained mobile or edge devices, where fast, on-device responses are increasingly in demand~\cite{10.1145/3534619, rashid2024tinyvqacompactmultimodaldeep,oh2025retabllama}. Furthermore, unless scaled to prohibitive sizes, these monolithic models can struggle with fine-grained details, often overlooking subtle but crucial information such as the units associated with numerical values~\cite{lin2022tsrformertablestructurerecognition,zhou2025enhancing}.

\subsection{Pipeline Approaches and Table Structure Recognition}
To address the efficiency limitations of large VLMs, a second stream of research has focused on pipeline-based approaches. A common pipeline involves using a model to first perform Optical Character Recognition (OCR) on the table image~\cite{cao2025tablemasterrecipeadvancetable,anand2024tcocr}, converting it into a structured format like HTML or Markdown, which is then fed to a Large Language Model (LLM) for reasoning. However, this conversion process is a major source of information loss. Crucial semantic information embedded in the visual layout, such as the relationships implied by merged cells or complex header structures, is often not fully captured in simple structured text. The importance of accurately parsing this visual layout has been highlighted by dedicated Table Structure Recognition (TSR) models like Table-Former~\cite{nassar2022tableformertablestructureunderstanding,kudale2025sprint,raja2020table,xue2021tgrnet,ngubane2024tableextractnet}, which aim to preserve this information. Even with perfect OCR, an LLM reasoning over a rigid, non-prose format can still fail to grasp the full context of the table. This has motivated a trend towards decomposing the vision and language tasks rather than treating them monolithically~\cite{liu2023deplotoneshotvisuallanguage}.

%% file: sections/method.tex
\section{Methodology}
We introduce {\name} (\textbf{T}able VQA via \textbf{A}ugmented \textbf{L}anguage-Enhanced \textbf{N}atural-text \textbf{T}ranscription), an LLM-centric framework for table question answering. Instead of relying on large end-to-end VLMs or brittle structured outputs, TALENT employs a lightweight VLM as a perception–narration module that produces two complementary views: (i) OCR text spans for symbolic precision, and (ii) a natural language narration that captures structural and semantic cues. These dual representations, combined with the question, are then passed to an LLM for reasoning and answer generation. Fig.~\ref{fig:methodology} illustrates the overall architecture of {\name}.

\subsection{Problem Formulation}
Let $T$ denote an input table image and $Q$ a natural language question about the table. The goal is to produce an answer $A$ in natural language:
\begin{equation}
    A = f(T, Q),
\end{equation}
where $f(\cdot)$ is the reasoning function implemented by the model. In standard Table VQA settings, $f$ is often realized either by an end-to-end VLM that maps directly from $(T, Q)$ to $A$, or by an OCR-to-LLM pipeline where $T$ is first converted into a structured representation (e.g., Markdown tables), and then sent to an LLM for question answering.

\subsection{Visual Information Extraction}

Given a table image $T$, the first stage of {\name} employs a lightweight VLM to extract two complementary forms of information: \emph{OCR spans} and a \emph{natural language narration}. This design stems from our core insight that OCR alone provides symbolic precision but often lacks the semantic richness needed for robust reasoning, while free-form narrations are expressive but may overlook exact details. By combining the two, we explicitly create a dual representation that balances precision and context.

\begin{figure}[htbp]
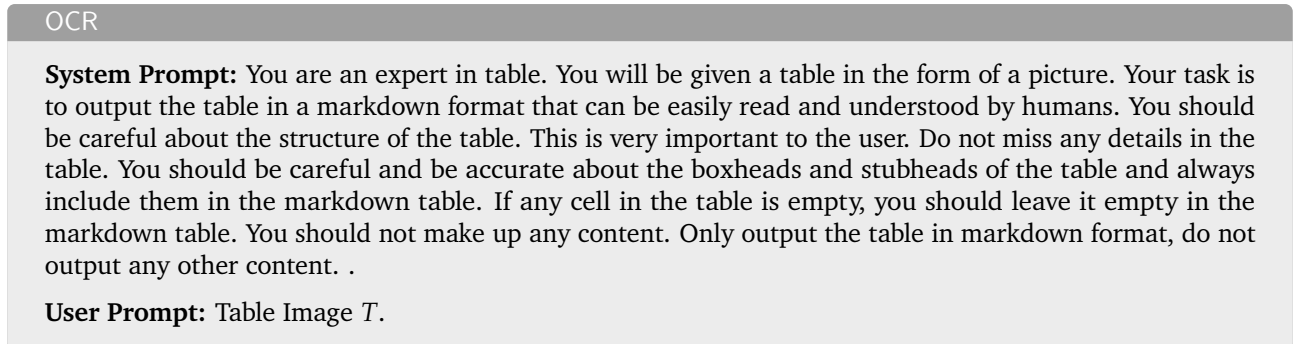

\centering
\begin{tcolorbox}[
    colback=gray!15,
    colframe=gray!75,
    title=OCR,
    fonttitle=\bfseries\sffamily\small\color{white},
    coltitle=white,
    bottomrule=0pt,
    toprule=0pt,
    leftrule=0pt,
    rightrule=0pt,
    rounded corners,
    width=1\linewidth
]
\small
\textbf{System Prompt:} You are an expert in table. You will be given a table in the form of a picture. 
Your task is to output the table in a markdown format that can be easily read and understood by humans.
You should be careful about the structure of the table. This is very important to 
the user. Do not miss any details in the table.
You should be careful and be accurate about the boxheads and stubheads of the 
table and always include them in the markdown table.
If any cell in the table is empty, you should leave it empty in the markdown table. 
You should not make up any content.
Only output the table in markdown format, do not output any other content.
.

\medskip
\textbf{User Prompt:} Table Image $T$.

\end{tcolorbox}
\caption{The prompt used for OCR.}
\label{fig:ocr}
% \vspace{-1.5em}
\end{figure}

\paragraph{OCR Spans.}  
We first obtain an OCR output $\mathcal{O}(T)$ that consists of recognized textual tokens:
\begin{equation}
    \mathcal{O}(T) = f_{\text{VLM}}(\texttt{Prompt}_{\text{OCR}}(T)),
\end{equation}
where $\mathcal{O}(T)$ will be structured output as markdown or HTML. We show the prompt used for OCR in Fig.~\ref{fig:ocr}. This symbolic representation provides high-precision textual content and preserves the literal strings from table cells. However, as our preliminary analysis showed, relying solely on $\mathcal{O}(T)$ often leads to reasoning errors because structural cues (e.g., merged headers or column dependencies) are not naturally represented.

\begin{figure}[htbp]
\centering
\begin{tcolorbox}[
    colback=gray!15,
    colframe=gray!75,
    title=Natural-text Transcription,
    fonttitle=\bfseries\sffamily\small\color{white},
    coltitle=white,
    bottomrule=0pt,
    toprule=0pt,
    leftrule=0pt,
    rightrule=0pt,
    rounded corners,
    width=1\linewidth
]
\small
\textbf{System Prompt:} You are an expert in table.
You will be given a table in the form of a picture.
Your task is to describe the content and structure of the table in details.
You should include any relevant table's details and context that may help the model understand the table.
This is very important to the user.

\medskip
\textbf{User Prompt:} Table Image $T$.

\end{tcolorbox}
\caption{The prompt used for Natural language narration.}
\label{fig:narration}
% \vspace{-1.5em}
\end{figure}

\paragraph{Natural Language Narration.}  
To address this limitation, we further prompt the same VLM to generate a natural language description 
\begin{equation}
    {N}(T)=f_{\text{VLM}}(\texttt{Prompt}_{\text{Narr}}(T))
\end{equation} 
that verbalizes the table, where the prompt used is shown in Fig.~\ref{fig:narration}. The narration highlights structural relations (e.g., "The first column lists the years, while the second column reports sales''), encodes contextual cues such as units or comparisons, and draws attention to salient entries. This component is motivated by the observation that LLMs are inherently optimized for natural language understanding; providing $\mathcal{N}(T)$ therefore allows the reasoning module to leverage its linguistic priors more effectively than with rigid structured formats like Markdown.

\paragraph{Dual Representation.}  
The outputs of this stage are thus a dual representation
\begin{equation}
    R(T) = \{\mathcal{O}(T), \mathcal{N}(T)\},
\end{equation}
which couples the \textbf{symbolic precision} of OCR with the \textbf{semantic richness} of narration.

\subsection{Language-based Question Answering}

The second stage of {\name} leverages an LLM as the primary reasoning engine. Given the dual representation $R(T) = \{\mathcal{O}(T), \mathcal{N}(T)\}$ and a question $Q$, the LLM generates the final answer $A$. Since LLMs are highly sensitive to instructions, we design a prompting strategy that explicitly guides the model to produce complete and unambiguous outputs. In particular, our approach is grounded in three key \emph{prompt design principles}:  

\paragraph{Explicit Unit Requirement.}  

Prompts are formulated to request answers that include both values and measurement units. For example, instead of "What is the revenue in Q3?", the prompt asks: "Please provide a complete sentence describing the revenue in Q3, including the amount and its unit." This mitigates the frequent omission of units in raw numerical responses.

\paragraph{Natural Language Format.}  

We encourage the model to answer in fluent, human-readable sentences rather than terse outputs. This format leverages the LLM's strength in generating coherent text and improves interpretability for end users.

\paragraph{Contextual Awareness.}  

Prompts highlight the importance of grounding answers in the contextual cues contained in $\mathcal{N}(T)$, ensuring that values are correctly associated with their headers and surrounding descriptions. This reduces ambiguity in complex table layouts.

Based on the three design principles, {\name} directly addresses prevalent issues in Table VQA such as unit omission, incomplete answers, and ambiguous expressions. By combining dual representations with carefully crafted prompts, the LLM can reliably reason over both the symbolic and semantic aspects of tables.

\section{ReTabVQA Dataset Curation}

While existing benchmarks such as TableVQA-Bench~\cite{kim2024tablevqabenchvisualquestionanswering} provide valuable evaluation resources, they primarily focus on direct value retrieval questions.  
To rigorously assess our proposed reasoning-oriented methods, we require a dataset that emphasizes multi-step quantitative reasoning over tabular data rather than simple lookup operations.  
Therefore, we curate a new benchmark, termed \textbf{ReTabVQA (Reasoning Table VQA)}, specifically designed to evaluate complex reasoning ability in table-based visual question answering.

The \textbf{ReTabVQA} dataset contains \textbf{120 question–answer pairs} derived from \textbf{60 table images} (15 images per category) across four distinct domains within the TableVQA-Bench~\cite{kim2024tablevqabenchvisualquestionanswering}  dataset.  
Each question is crafted to require reasoning across multiple data points, integrating arithmetic, comparison, and synthesis operations.

\begin{figure*}[h!]
    \centering
    \includegraphics[width=0.8\columnwidth]{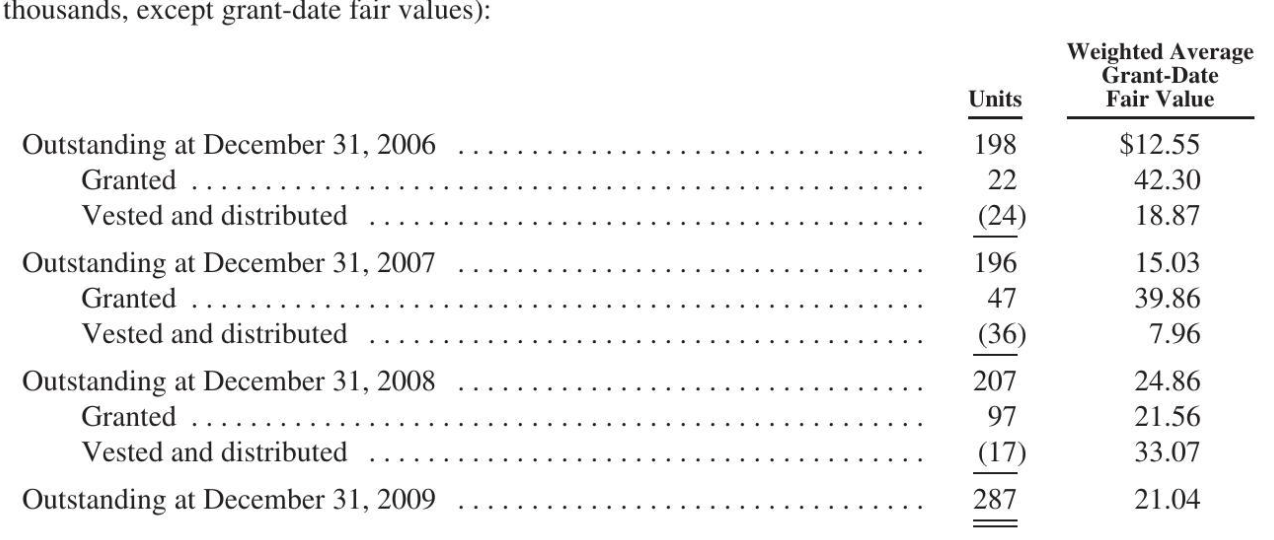}
    \caption{Example Table from the ReTabVQA Dataset.}
    \label{fig:example_table}
    % \vspace{-1.5em}
\end{figure*}

For instance, the original TableVQA-Bench question for the table shown in Figure~\ref{fig:example_table} is:  
\textit{``How many units were outstanding at December 31, 2007?''}, with the answer \textbf{196}.  

In contrast, our \textbf{ReTabVQA} dataset introduces two reasoning-oriented questions for the same table:  
(1) \textit{``What is the percentage change in the total fair value of outstanding units from 2006 to 2007?''}, where the total fair value increases from $198\times12.55=2{,}484.9$k to $196\times15.03=2{,}945.9$k, resulting in an \textbf{18.55\%} increase; and  
(2) \textit{``For 2009, what was the net change in total fair value of units resulting from granted and vested/distributed activities?''}, computed as $(97\times21.56)-(17\times33.07)=\textbf{1{,}529.13}$k.  

These examples highlight that \textbf{ReTabVQA} emphasizes compositional and multi-step quantitative reasoning, moving beyond direct data retrieval.

% Optional dataset summary table (you can comment out if not needed)
\begin{table}[h!]
\centering
\caption{Statistics of the ReTabVQA Dataset.}
\label{tab:retabvqa_stats}
\vspace{0.3em}
\begin{tabular}{lcc}
\toprule
\textbf{Category} & \textbf{\# Tables} & \textbf{\# QA Pairs} \\
\midrule
Financial Reports & 15 & 30 \\
Sports Statistics & 15 & 30 \\
Survey Results & 15 & 30 \\
Scientific Tables & 15 & 30 \\
\midrule
\textbf{Total} & \textbf{60} & \textbf{120} \\
\bottomrule
\end{tabular}
\vspace{-0.5em}
\end{table}

%% file: sections/experiments.tex
\section{Experiment}
In the experiments, we evaluate {\name} and answer the following questions: RQ 1)~How does {\name} compare to state-of-the-art (SOTA) Table VQA methods? RQ 2)~How do VLM and LLM components contribute to the overall performance in the proposed {\name}? RQ 3)~How does the size of VLMs and LLMs used in {\name} influence the performance?

% \begin{itemize}
% \item \textbf{RQ1:} How does {\name} compare to state-of-the-art (SOTA) Table VQA methods?

% \item \textbf{RQ2:} How do VLM and LLM components contribute to the overall performance in the proposed {\name}?
% \item \textbf{RQ3:} How does the size of VLMs and LLMs used in {\name} influence the performance?
% \end{itemize}

\subsection{Dataset and Experimental Setup}

\subsubsection{Dataset and Evaluation Metric}
We conduct experiments on the public \href{https://huggingface.co/datasets/terryoo/TableVQA-Bench}{TableVQA-Bench} dataset~\cite{kim2024tablevqabenchvisualquestionanswering}, a comprehensive benchmark for Table VQA. This dataset contains a diverse collection of table images and corresponding questions across multiple domains, with varying levels of complexity. Ground-truth answers are annotated with numerical values and units, providing a solid foundation for evaluating unit-aware accuracy and contextual completeness. 

In addition, we evaluate our method on the newly constructed \textbf{ReTabVQA Dataset }, which extends TableVQA-Bench with multi-step, numerically grounded reasoning questions. This dataset is specifically designed to assess a model’s ability to synthesize information across multiple table entries and perform compositional quantitative reasoning.

Following~\cite{kim2024tablevqabenchvisualquestionanswering}, we use accuracy as the evaluation metric. A prediction is considered correct if it contains the ground-truth answer.

\subsubsection{Baselines}
We compare {\name} against the following baselines: 1)~Direct Prompt~\cite{kim2024tablevqabenchvisualquestionanswering}, 2)~Perfect OCR, 3)~Generated OCR~\cite{kim2024tablevqabenchvisualquestionanswering}, and 4)~Language Description. \textbf{Direct Prompt} is a classic approach that asks a VLM to generate an answer directly from the image and question, without relying on any LLM.  
\textbf{Perfect OCR} provides the LLM with ground-truth OCR outputs of the table image, followed by the question, to evaluate whether the LLM can understand the OCR structure and answer accordingly.  
\textbf{Generated OCR} involves prompting a VLM to generate the OCR results from the table image, which are then fed into an LLM to produce the final answer.  
\textbf{Language Description} prompts the VLM to first generate a detailed description of the table content and then generate the final answer based on this description.

\subsubsection{Implementation Details}
Our experiments use the Qwen2.5 model series as backbones, including Qwen2.5-VL-3B~\cite{bai2025qwen25vltechnicalreport}, Qwen2.5-VL-7B~\cite{bai2025qwen25vltechnicalreport}, Qwen2.5-3B~\cite{qwen2025qwen25technicalreport}, and Qwen2.5-7B~\cite{qwen2025qwen25technicalreport}. All models are deployed with BF16 precision using vLLM (v0.9.2)~\cite{vllm2025supported} and FlashAttention-2~\cite{dao2023flashattention2fasterattentionbetter}. We set the inference parameters to temperature = 0.1 and top\_p = 0.9. We also include several SOTA VLMs as baselines: Phi-4-multimodal (14B)~\cite{microsoft2025phi4minitechnicalreportcompact}, MiniCPM (3B)~\cite{hu2024minicpmunveilingpotentialsmall}, GPT-4V~\cite{2023GPT4VisionSC}, GPT-4~\cite{openai2024gpt4technicalreport}, Gemini-ProV~\cite{geminiteam2025geminifamilyhighlycapable}, and Gemini-Pro~\cite{geminiteam2025geminifamilyhighlycapable}.

\subsection{Performance Comparison}

\begin{table}[t]
\centering
\caption{Performance of {\name} and baselines. Results of $^{\star}$ are taken from \cite{kim2024tablevqabenchvisualquestionanswering}.}
\label{tab:main_results}
% \resizebox{0.5\textwidth}{!}{
\begin{tabular}{l c c c}
\toprule
\textbf{Method} & \textbf{VLM} & \textbf{LLM} &  \textbf{Accuracy (\%)} \\
\midrule
\multicolumn{4}{l}{\textit{\textbf{Baseline Methods}}} \\
\midrule
Direct Prompt$^{\star}$ & GPT-4V &--- &54.50\\
Direct Prompt$^{\star}$ & Gemini-ProV &--- &38.30\\
Direct Prompt   & MiniCPM (3B)             & ---          & 74.20 \\
Direct Prompt   & 
Phi-4-multimodal (14B)             & ---            & 66.80 \\
\midrule
Perfect OCR      & ---              & Qwen2.5-3B   & 72.93 \\
Perfect OCR      & ---              & Qwen2.5-7B   & 75.47 \\
Perfect OCR   & ---              & Qwen3-235B            & 80.40 \\
\midrule
VLM Generated  OCR   & Qwen2.5-VL-3B              & Qwen2.5-3B   & 71.47  \\
VLM Generated  OCR & Qwen2.5-VL-7B & Qwen2.5-7B  & 73.60 \\
\midrule
Language Description& Qwen2.5-VL-3B       & Qwen2.5-3B                 & 68.07 \\
Language Description& Qwen2.5-VL-7B     & Qwen2.5-7B                 & 72.27 \\

%Direct Prompt   & QVQ-Max              &  ---          & 217 & \textbf{85.53} \\
\midrule
Generated OCR$^{\star}$   & GPT-4V            & GPT-4       & 60.70\\
Generated OCR$^{\star}$   & Gemini-ProV            & Gemini-Pro  & 48.60\\
\midrule
\multicolumn{4}{l}{\textit{\textbf{TALENT (Our Proposed Method)}}} \\
\midrule
{\name} & Qwen2.5-VL-3B& Qwen2.5-3B  & 74.73 \\
{\name} & Qwen2.5-VL-7B & Qwen2.5-3B  & 76.53 \\
{\name} & Qwen2.5-VL-7B & Qwen2.5-7B   & 80.67 \\
{\name} & Qwen2.5-VL-3B & Qwen2.5-7B  & \textbf{81.13} \\
%TALENT & QVQ      & Qwen2.5-7B    & 137* & 84.78 \\
\bottomrule
\end{tabular}
% } 
\vspace{-1em}
\end{table}

In this section, we compare {\name} with various baselines to answer RQ1 and RQ2, shown in Tab.~\ref{tab:main_results}. We observe the following key findings:

First, {\name} consistently outperforms all baseline methods across different model combinations. The best-performing {\name} variant, using Qwen2.5-VL-3B as the VLM and Qwen2.5-7B as the LLM, achieves an accuracy of 81.13\%, which surpasses even the strongest baseline, Perfect OCR with Qwen3-235B-A22B (thinking mode) (80.40\%). This demonstrates the effectiveness of {\name}'s two-stage pipeline that integrates both OCR understanding and language-based reasoning.

Second, methods that rely solely on generated OCR exhibit significantly lower performance, even when paired with powerful language models such as GPT-4 or Gemini-Pro. For example, the Generated OCR baseline achieves only 60.70\% with GPT-4 and 48.60\% with Gemini-Pro. This suggests that OCR results produced by vision-language models are often noisy and difficult for LLMs to interpret accurately, limiting their overall performance.

Third, even with perfect OCR inputs, baseline methods still struggle to achieve high accuracy, with results ranging from 72.93\% to 80.40\%. This indicates that language models alone have difficulty reasoning over structured OCR outputs, highlighting the importance of transforming table content into interpretable natural language, a core component of the {\name} design.

Finally, natural language description alone is also insufficient. The Description Prompt baseline, which uses only textual summaries without incorporating OCR outputs, performs worse than {\name}, with accuracies of 68.07\% and 72.27\%. This confirms that relying solely on high-level descriptions limits the model’s ability to reason about fine-grained table content, further validating {\name}’s design that integrates both structured OCR results and natural language understanding.

\subsection{ReTabVQA Results and Analysis}

\begin{table}[h!]
\centering
\small
\caption{Performance comparison of different models on the \textbf{ReTabVQA Dataset}. All models are from the Qwen2.5 series, and the model configuration is denoted as (VLM size - LLM size).}
\label{tab:new_dataset_results}
% \resizebox{1\columnwidth}{!}{
\begin{tabular}{lcc}
\toprule
\textbf{Approach} & \textbf{Model Configuration} & \textbf{Accuracy (\%)} \\
\midrule
\multirow{3}{*}{Generated OCR} & \hspace{1em} 3B-3B & 47.80 \\
                               & \hspace{1em} 7B-7B & 46.51 \\
                               & \hspace{1em} 7B-3B & 46.98 \\
\midrule
\multirow{3}{*}{\textbf{TALENT (Ours)}} & \hspace{1em} 3B-3B & \textbf{52.50} \\
                                     & \hspace{1em} 7B-7B & \textbf{55.06} \\
                                     & \hspace{1em} 7B-3B & \textbf{55.00} \\
\bottomrule
\end{tabular}
% }
% \vspace{-1.6em}
\end{table}

\noindent\textbf{Experimental Results and Analysis.}  
We evaluated TALENT against a baseline on the \textbf{ReTabVQA Dataset} using three model configurations for fair comparison. The baseline adopts standard prompting, while TALENT employs our full dual-representation and human-language prompting strategy.  

As shown in Table~\ref{tab:new_dataset_results}, TALENT consistently outperforms the baseline across all models. For \texttt{3B-3B}, TALENT achieves \textbf{52.50\%} vs. \textbf{47.80\%}; for \texttt{7B-7B}, \textbf{55.06\%} vs. \textbf{46.51\%}; and for \texttt{7B-3B}, \textbf{55.00\%} vs. \textbf{46.98\%}. These results confirm the effectiveness of our dual-representation strategy, yielding consistent gains across model sizes and highlighting that the TALENT framework itself—rather than specific model capacity—is key to improved Table VQA performance.  

\noindent\textbf{Error Analysis.}  
Overall accuracy is lower on the more challenging \textbf{ReTabVQA Dataset}. Most errors occur not in data retrieval but in subsequent numerical calculations, indicating that the mathematical reasoning of smaller LLMs remains a bottleneck even with accurate contextual inputs. These results further demonstrate the robustness and generalization of TALENT in complex table reasoning scenarios.

\subsection{Impact of Model Scaling}

To examine how the capacities of the visual-language model (VLM) and the language model (LLM) jointly affect performance, we conduct a scaling analysis using the Qwen2.5 series with model sizes of 3B, 7B, and 32B. Table~\ref{tab:scaling_matrix_3x3} reports the accuracy (\%) under all combinations of VLM and LLM sizes.

We fit the empirical results using several candidate functions, including log-linear, power-law, and saturating (logistic) models. The log-linear form provides the most stable and interpretable fit, with a coefficient of determination $R^{2}=0.83$:
\begin{equation}
    A \approx 73.01 + 0.84 \log(S_V) + 2.66 \log(S_L)
    \label{eq:scaling_model}
\end{equation}
where $A$ denotes accuracy, and $S_L$ and $S_V$ represent the parameter sizes of the LLM and VLM, respectively. Both components contribute positively but sublinearly to accuracy, showing diminishing returns with scale. The LLM coefficient (2.66) notably exceeds that of the VLM (0.84), suggesting that scaling the language component yields a greater performance gain.

\noindent\textbf{Effect of LLM Scaling.}  
Holding the VLM fixed at 3B, enlarging the LLM from 3B to 7B improves accuracy by \textbf{+6.4} points (74.73\% $\rightarrow$ 81.13\%), and a further increase to 32B yields an additional \textbf{+2.6} points (83.73\%). This pattern aligns with the regression coefficient $\beta_L = 2.66$, indicating that LLM capacity is the primary driver of reasoning improvement.

\noindent\textbf{Effect of VLM Scaling.}  
When fixing the LLM at 3B, scaling the VLM from 3B to 7B and 32B enhances accuracy by \textbf{+1.8} and \textbf{+3.1} points, respectively. Although visual encoding capacity contributes positively, its influence ($\beta_V = 0.84$) is substantially smaller than that of the LLM.

\noindent\textbf{Conclusion.}  
The regression with $R^{2}=0.83$ confirms a near-logarithmic scaling trend across both modalities. The contribution of the LLM is approximately \textbf{three times stronger} than that of the VLM ($\beta_L / \beta_V \approx 3.2$), supporting an \textit{LLM-centric design}: once the VLM provides sufficiently informative visual representations, the reasoning capability and scale of the LLM dominate performance in Table VQA.

\begin{table}[h!]
\centering
\caption{Accuracy (\%) comparison for different VLM and LLM sizes from the Qwen2.5 series, including 32B models.}
\label{tab:scaling_matrix_3x3}
\begin{tabular}{l|ccc}
\toprule
\diagbox[width=8em]{VLM Size}{LLM Size} & \textbf{3B} & \textbf{7B} & \textbf{32B} \\
\midrule
\textbf{3B} & 74.73 & 81.13 & 83.73 \\
\textbf{7B} & 76.53 & 80.67 & 83.25 \\
\textbf{32B} & 79.60 & 81.95 & \textbf{83.80} \\
\bottomrule
\end{tabular}
\end{table}

\begin{figure}[h!]
    \centering
    \includegraphics[width=0.6\columnwidth]{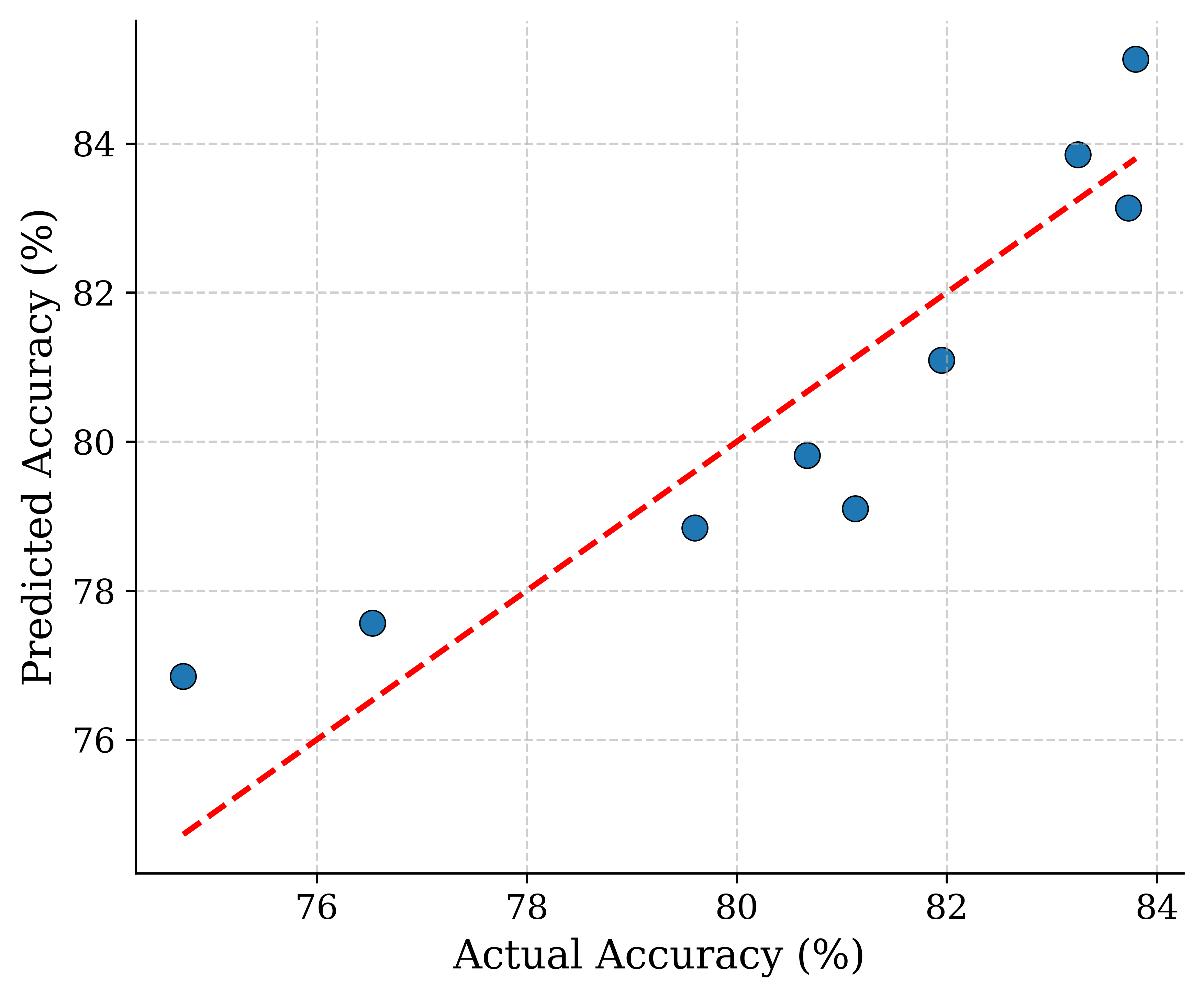}
    \caption{Log-linear regression fit of model scaling based on Equation~\ref{eq:scaling_model}. Each point represents a configuration of VLM and LLM sizes, comparing the \textit{actual} accuracy from experiments with the \textit{predicted} accuracy computed by the fitted model. The close alignment along the diagonal ($R^{2}=0.83$) confirms that the log-linear scaling model accurately captures the relationship between model size and performance, with a stronger effect from LLM scaling.}
    \label{fig:scaling_fit}
    % \vspace{-1.6em}
\end{figure}

\subsection{Impact of Input Image Resolution}

To investigate the effect of input image resolution on performance, we conducted a further analysis using the \texttt{Qwen2.5-vl-3B to Qwen2.5-3B} model configuration. We tested both our TALENT method and the baseline ``Generated OCR'' approach at two different input resolutions: 512x512 and 1024x1024. The results are summarized in Table~\ref{tab:resolution_study}.

\begin{table}[h!]
\centering
\small
\caption{Impact of input image resolution on accuracy for the Qwen2.5-vl-3B to Qwen2.5-3B Model Configuration .}
\label{tab:resolution_study}
\begin{tabular}{llcc}
\toprule
\textbf{Method} & \textbf{Resolution} & \textbf{Accuracy (\%)} \\
\midrule
\multirow{2}{*}{Generated OCR}       & 512$\times$512 & 69.54 \\
                                     & 1024$\times$1024  & 71.47 \\
\midrule
\multirow{2}{*}{\textbf{TALENT~(Ours)}} & 512$\times$512  & 73.00 \\
                                     & 1024$\times$1024  & \textbf{74.73} \\                                    
\bottomrule
\end{tabular}
% \vspace{-1.5em}
\end{table}

\begin{figure*}[h!]
    \centering
    \includegraphics[width=0.95\textwidth]{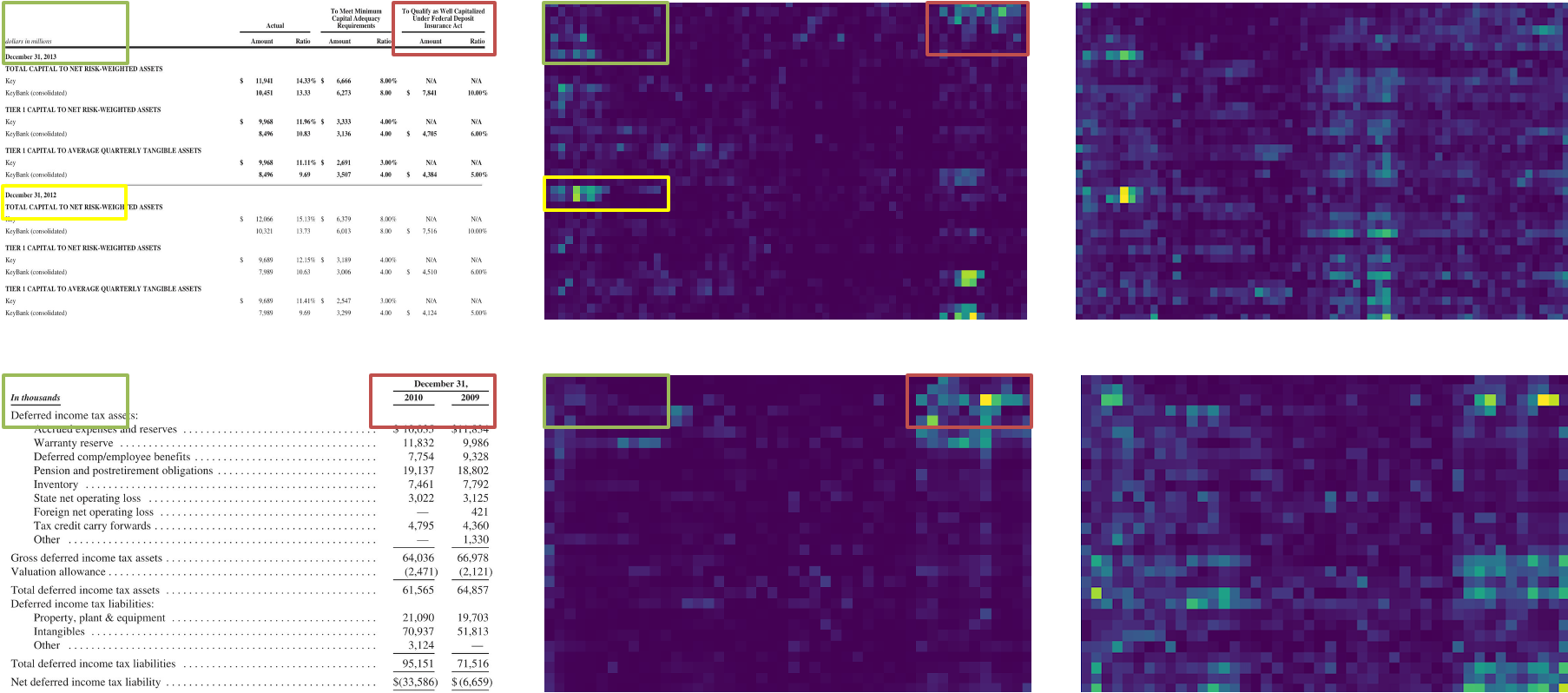}
    \caption{Two case studies from the TableVQA-Bench dataset. \textbf{Top Row:} A complex table where the unit "(In millions)" is in the header. \textbf{Bottom Row:} A simpler table where the unit "(in thousands)" is in the top-left corner. In both cases, the attention map for direct prompting (Second column) is sparse, while the map for generating a description (Third column) is more holistic, leading to correct unit interpretation. }
    \label{fig:case_study}
\end{figure*}

We conclude two findings. First, increasing input resolution from 512 to 1024 improves performance for both methods. TALENT’s accuracy rises by \textbf{1.73} points (73.00\% → 74.73\%), while the baseline gains \textbf{1.93} points, indicating that higher resolution yields finer visual details beneficial to data extraction. This may result from the Qwen-VL series’ shared ViT vision encoder and LM decoder architecture, which is sensitive to input resolution.

%\textbf{This positive correlation also indicates that the Qwen series models are sensitive to input resolution.}

Secondly, TALENT framework consistently outperforms the ``Generated OCR'' baseline at both resolutions. At the higher 1024 resolution, TALENT maintains a significant lead of more than 3 percentage points. This indicates that while higher resolution is beneficial for all methods, our dual-representation strategy provides a robust advantage that is independent of the input resolution.

\subsection{Case Study}

To further illustrate the effectiveness of our proposed natural language narration prompting, we present a case study from the TableVQA-Bench dataset (Figure~\ref{fig:case_study}). This example highlights common pitfalls of direct VLM responses and how our method mitigates the unit omission problem through table-wide contextual understanding.

\textbf{Scenario 1: Direct VLM Response.}
\noindent When directly prompted, baseline VLMs often fail to interpret contextual units. In the top example of Figure~\ref{fig:case_study}, when asked for the "well capitalized" amount for KeyBank, the model outputs "4,124" but ignores the \textbf{"(In millions)"} unit in the header. Similarly, in the lower table, when asked for the "Warranty reserve" value, the model answers "11,832" without the \textbf{"(in thousands)"} unit in the corner. As shown by its sparse attention map (Figure~\ref{fig:case_study}, second column), the model narrowly focuses on target cells, neglecting global context—leading to answers off by several orders of magnitude.

\textbf{Scenario 2: Our Proposed Method ({\name}).}
\noindent In contrast, our two-stage method first builds a holistic understanding of the table. The descriptive task drives the VLM to scan the entire layout (Figure~\ref{fig:case_study}, third column), successfully capturing both "(In millions)" and "(in thousands)" unit declarations. The resulting context is then passed to the LLM, which, guided by our language prompting strategy, generates complete, unambiguous responses: \textbf{"The amount to qualify as well capitalized was \$4,124 million"} and \textbf{"The value of the Warranty reserve on December 31, 2010, was \$11,832,000."} This demonstrates that our description-driven approach effectively handles diverse unit placements and complex layouts, resolving the unit omission issue that hampers direct prompting.

%% file: sections/conclusion.tex
\section{Conclusion}
\label{sec:conclusion}

This paper presents {\name}, a framework that redefines the roles of VLMs and LLMs in Table VQA.  
Instead of relying on monolithic end-to-end VLMs or fragile structured representations, {\name} employs a lightweight VLM as a perception and narration module to produce dual representations: OCR spans for symbolic precision and natural language narration for semantic context. These representations are fused with the question for reasoning by an LLM.  
Extensive experiments lead to three main insights.  
First, {\name} consistently surpasses strong baselines across different backbones, showing that compact VLM–LLM pairs can rival or exceed much larger models.  
Second, ablation studies verify the complementary roles of OCR spans and natural language narration, as removing either substantially reduces performance.  
Third, scaling analysis indicates that while both components benefit from larger capacities, the LLM contributes more significantly, validating our LLM-centric design principle.  
Overall, {\name} provides an efficient and accurate solution for Table VQA, integrating symbolic grounding with natural language reasoning and enabling deployment in resource-limited environments such as mobile and edge devices.